\documentclass[runningheads]{llncs}

\usepackage{eccv}

\usepackage{eccvabbrv}

\usepackage{graphicx}
\usepackage{booktabs}

\usepackage[accsupp]{axessibility}  

\usepackage{hyperref}

\usepackage{orcidlink}

\usepackage{booktabs} 

\usepackage{url}
\usepackage{lipsum}
\usepackage{xcolor}
\usepackage{bbding}
\usepackage{bm}
\usepackage{enumitem}
\usepackage{multirow}
\usepackage{xspace}

\usepackage{colortbl}
\usepackage{booktabs}
\usepackage{tabularx}

\usepackage{wrapfig}

\newcommand\mypara[1]{\vspace{5pt}\noindent\textbf{#1.}}

\begin{document}

\title{StoryBlender: Inter-Shot Consistent and Editable 3D Storyboard with Spatial-temporal Dynamics} 

\titlerunning{StoryBlender: Consistent 3D Storyboard Generation}

\author{Bingliang Li\inst{1}\thanks{Equal Contribution}\orcidlink{0009-0002-0774-9308} \and
Zhenhong Sun\inst{2}{$^\star$}\orcidlink{0009-0000-5135-0658} \and
Jiaming Bian\inst{3}\orcidlink{0009-0004-9377-3420} \and
Yuehao Wu\inst{1}\orcidlink{0009-0003-1446-9088} \and
Yifu Wang\inst{4}\orcidlink{0000-0002-7901-483X} \and
Hongdong Li\inst{2}\orcidlink{0000-0003-4125-1554} \and
Yatao Bian\inst{5}\orcidlink{0000-0002-2368-4084} \and
Huadong Mo\inst{1}\thanks{Corresponding author}\orcidlink{0000-0002-7782-2884} \and
Daoyi Dong\inst{6}\orcidlink{0000-0002-7425-3559}
}

\authorrunning{B.~Li et al.}

\institute{University of New South Wales, Australia.
\email{lbingl@outlook.com, \{yuehao.wu,huadong.mo\}@unsw.edu.au} \and
Australian National University, Australia.
\email{\{zhenhong.sun,hongdong.li\}@anu.edu.au} \and
Central South University, China.
\email{bianjiaming@csu.edu.cn} \and
Vertex Lab, China.
\email{usasuper@126.com} \and
National University of Singapore, Singapore.
\email{ybian@nus.edu.sg} \and
University of Technology Sydney, Australia.
\email{daoyi.dong@uts.edu.au}
}

\maketitle

\begin{abstract}
Storyboarding is a core skill in visual storytelling for film, animation, and games. However, automating this process requires a system to achieve two properties that current approaches rarely satisfy simultaneously: \emph{inter-shot consistency} and \emph{explicit editability}. While 2D diffusion-based generators produce vivid imagery, they often suffer from identity drift along with limited geometric control; conversely, traditional 3D animation workflows are consistent and editable but require expert-heavy, labor-intensive authoring.  We present \textbf{StoryBlender}, a grounded 3D storyboard generation framework governed by a Story-centric Reflection Scheme. At its core,  we propose {\em StoryBlender} system, which is built on a three-stage pipeline: (1) Semantic-Spatial Grounding, to construct a continuity memory graph to decouple global assets from shot-specific variables for long-horizon consistency; (2) Canonical Asset Materialization, to instantiate entities in a unified coordinate space to maintain visual identity; and (3) Spatial-Temporal Dynamics, to achieve layout design and cinematic evolution through visual metrics. By orchestrating multiple agents in a hierarchical manner within a verification loop, StoryBlender iteratively self-corrects spatial hallucinations via engine-verified feedback. The resulting native 3D scenes support direct, precise editing of cameras and visual assets while preserving unwavering multi-shot continuity. Experiments demonstrate that StoryBlender significantly improves consistency and editability over both diffusion-based and 3D-grounded baselines. Code and dataset are available at \href{https://engineeringai-lab.github.io/StoryBlender/}{https://engineeringai-lab.github.io/StoryBlender.}


\end{abstract}

\begin{figure}
    \centering
    \includegraphics[width=\linewidth]{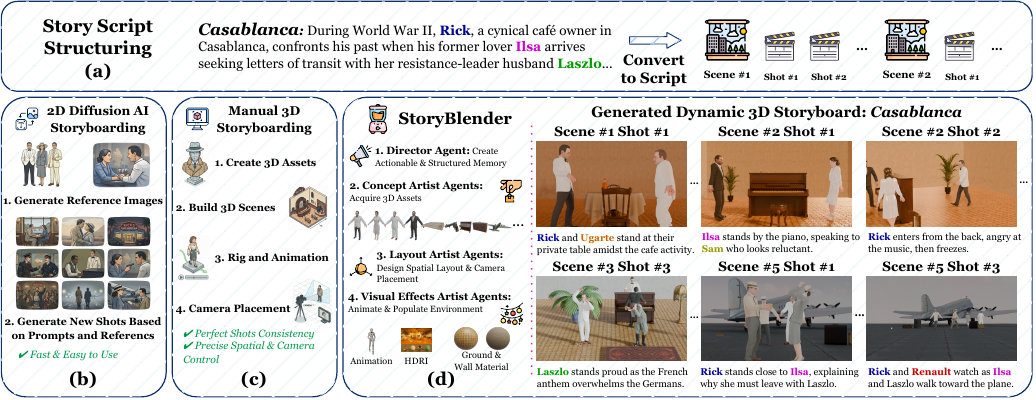}
    \caption{\textbf{Overview of StoryBlender compared to existing storyboarding methods.}
    \textbf{Left:} diffusion-based generation in pixel space; improving consistency typically requires reference inputs. 
    \textbf{Middle:} traditional 3D workflow; strong control but a complex, labor-intensive pipeline. 
    \textbf{Right (ours):} StoryBlender uses a hierarchical multi-agent planning framework to create consistent, editable 3D storyboards across shots.}
    \label{fig:paradigm}
\end{figure}

\section{Introduction}

Visual storytelling aims to translate a narrative (e.g., a script or outline) into a sequence of coherent visual shots.
In practice, this is realized through storyboarding, an explicit, shot-by-shot plan that externalizes key creative decisions (characters, actions, layout, and camera intent) to coordinate teams and de-risk production during pre-production and pre-visualization in film, animation, and games.
At its essence, storyboarding demands two key capabilities: \textbf{consistency} and \textbf{editability}. Consistency means maintaining character identity, visual style, and 3D scene state (layout, props, lighting, etc.) across shots and largely determines whether the storyboard is practically usable. Editability means these elements are explicitly controllable and revisable (e.g., adjusting blocking) without redoing the entire sequence, and determines the practical usability of the system.

Existing solutions for automated storyboarding can be broadly grouped into two paradigms that trade off these two requirements: (i) 2D diffusion-based generation can render vivid, high-quality shots~\cite{zhou2024storydiffusion, dinkevich2025Story2Board,zheng2025fairygenstoriedcartoonvideo,zhang2025storymemmultishotlongvideo,tewel2024trainingfreeconsistenttexttoimagegeneration,11130395,huang2024incontextloradiffusiontransformers,shi2025AniMaker}, but achieving shot-to-shot consistency is difficult and often depends on reference inputs, while the pixel-space representation provides limited editability for geometric/cinematic changes. (ii) Traditional 3D workflows can provide explicit scene representations that naturally support consistency and editability, but they typically require a complex pipeline and substantial manual labor from skilled professionals, making the workload prohibitive at scale. This motivates using Large Language Model (LLM) agents to translate scripts into structured 3D actions (asset selection, placement, camera design) and automate expert-heavy steps while retaining controllability~\cite{huang2024toward, hu2024scenecraft, liu2025worldcraft, lin2025pat3d, huang2024story3d,sun20253dgeneralist}. However, LLMs alone cannot solve the entire pipeline end-to-end. They fundamentally lack a persistent, grounded world model to bridge the gap between semantic narrative planning and precise geometric execution. Specifically, LLMs often exhibit spatial hallucinations, proposing layouts or actions that conflict with physical constraints, and they lack an internal verification loop to correct these geometric errors. Moreover, maintaining narrative coherence requires tracking an evolving world state beyond the limited context window of a standard LLM.
This raises a question: \textit{Do general-purpose LLMs truly understand 3D geometry and spatial information to ensure inter-shot consistency and editability, or do they merely hallucinate semantic plausibility without using spatial grounding?}

To answer this, we introduce StoryBlender, a novel framework that reformulates storyboard generation not as stochastic image synthesis but as a Hierarchical Multi-Agent Planning process governed by a Story-centric Reflection Scheme. Unlike prior art, StoryBlender treats the narrative as a persistent, grounded simulation where agents collaborate within a 3D engine to ensure the output is both inter-shot consistent and editable. Our methodology is organized into three progressive stages that systematically operationalize the spatial-temporal dynamics of the story:
First, to establish a world model that transcends the limited context windows of LLMs, we propose Semantic-Spatial Grounding. Here, a Director Agent decomposes the narrative into a structured continuity memory graph, serving as a digital script supervisor that ensures precise information flow and maintains consistency across long horizons. Second, to eliminate generative hallucinations, Canonical Asset Materialization instantiates entities from the memory graph into a unified coordinate space, ensuring that character identities and props remain visually identical throughout the sequence. Finally, Spatial-Temporal Dynamics orchestrates the physical arrangement and cinematic evolution of these assets. By employing discrete visual servoing and programmatic engine verification, this stage resolves the complex dynamics of characters, cameras, and visual effects.
By integrating this persistent memory with a closed-loop reflection system, StoryBlender replaces mere semantic plausibility with rigorous physical grounding. This architecture also ensures that every creative decision remains explicitly editable within the 3D environment, providing a robust solution for consistent 3D storyboarding.

The main contributions are summarized as follows:
\begin{itemize}[leftmargin=*, noitemsep, nolistsep]
    \item[$\bullet$] We reformulate 3D storyboarding as a closed-loop optimization process by pairing LLM reasoning with in-engine physical verification to iteratively self-correct spatial hallucinations.
    \item[$\bullet$] We introduce a structured hierarchical memory to decouple global assets from shot-specific variables, enabling seamless asset materialization and ensuring long-horizon identity consistency across the entire narrative.
    \item[$\bullet$] We propose a modular pipeline that resolves complex layout and cinematic evolution through visual metrics, achieving a storyboard representation that is both physically grounded and explicitly editable.
\end{itemize}

\section{Related Work}

\mypara{Generative Visual Storytelling}
Recent progress has pushed visual storytelling from single images to coherent narrative sequences, driven by the emergence of powerful video foundation models such as Stable Video Diffusion~\cite{blattmann2023stable} and Sora~\cite{liu2024sora}. To adapt these models for narrative control, prior work improves character consistency with visual adapters (e.g., IP-Adapter~\cite{ye2023ip}, T2I-Adapter~\cite{mou2024t2i}, StoryAdapter~\cite{mao2024story}) to strengthen cross-frame alignment through mechanisms such as consistent attention and latent anchoring (e.g., StoryDiffusion~\cite{zhou2024storydiffusion}, StoryGen~\cite{liu2024intelligent}, Story2Board~\cite{dinkevich2025Story2Board}). For multi-shot narratives, agent-style pipelines increasingly use LLMs to plan semantic layouts and orchestrate the generation process step-by-step (e.g., VideoDirectorGPT~\cite{lin2023videodirectorgpt}, MovieAgent~\cite{wu2025automated}, DreamFactory~\cite{xie2024dreamfactory}, AniMaker~\cite{shi2025AniMaker}, CineVision~\cite{wei2025cinevision}). Recent diffusion-control studies further indicate that reliable complex visual generation often requires explicit instance-, sketch-, and motion-level constraints~\cite{yao2023dance,sun2024eggen,sun2024t,zhang2025hierarchical}. However, because these 2D-centric methods operate entirely in pixel or latent space, they fundamentally lack an underlying physical model. This inherently leaves them prone to geometric hallucinations, viewpoint inconsistencies, and weak environmental persistence during editing, limiting their reliability for rigorous pre-visualization workflows.

\mypara{LLM-Driven 3D Scene Synthesis}
To gain spatial consistency, recent systems couple LLMs with 3D representations to generate scenes through scripting or layout planning (e.g., SceneCraft~\cite{hu2024scenecraft}, I-Design~\cite{celen2024idesign}, SceneWeaver~\cite{yang2025sceneweaver}, WorldCraft~\cite{liu2025worldcraft}, PAT3D~\cite{lin2025pat3d}, Physcene~\cite{yang2024physcene}). In these pipelines, the LLM typically acts as a high-level planner, decomposing text into object inventories and spatial relations. Recent advancements have further integrated LLMs directly into the generation pipeline using diverse strategies. For instance, SceneTeller~\cite{ocal2024sceneteller} predicts 3D layouts via in-context learning. Other approaches leverage Chain-of-Thought for direct numerical spatial reasoning~\cite{ran2025direct}, or connect multi-agent frameworks to professional game engines for procedural generation (UnrealLLM~\cite{songtang2025unrealllm}). Yet most treat generation as a one-off, \textit{static} arrangement problem, optimizing a single snapshot without mechanisms to maintain state across shots or to reuse scene constraints (e.g., persistent props, character interactions, and camera settings). Reliable character interaction and motion grounding also motivate open-world HOI detection and real-world 3D human pose benchmarking~\cite{yang2024open,wang2024freeman}.

\mypara{3D Cinematography and Pre-visualization}
Engine-based pre-visualization methods, such as Virtual Dynamic Storyboard (VDS)~\cite{rao2023dynamic}, often rely on pre-selected assets and manual setup. Recently, AI-driven approaches have sought to automate and improve this pipeline. Generative methods are increasingly integrating with 3D blocking to provide robust cinematic control, such as CinePreGen~\cite{chen2024cinepregen}, which utilizes engine-powered diffusion for camera-controllable pre-visualization, and PrevizWhiz~\cite{hu2026previzwhiz}, which combines rough 3D scenes with 2D references. Other works explore the transformation of 2D sketch storyboards into 3D animations (Sketch2Anim~\cite{zhong2025sketch2anim}). Story3D-Agent~\cite{huang2024story3d} scripts 3D actions with LLMs as linear sequences in a pre-configured environment. Despite these advances, existing systems lack a global context manager for multi-scene transitions. StoryBlender addresses these gaps via a continuity memory graph that maintains persistent states and enforces inter-shot consistency.

\section{Methodology}

\subsection{Overview}
\begin{figure}[t]
  \centering
  \includegraphics[width=\textwidth]{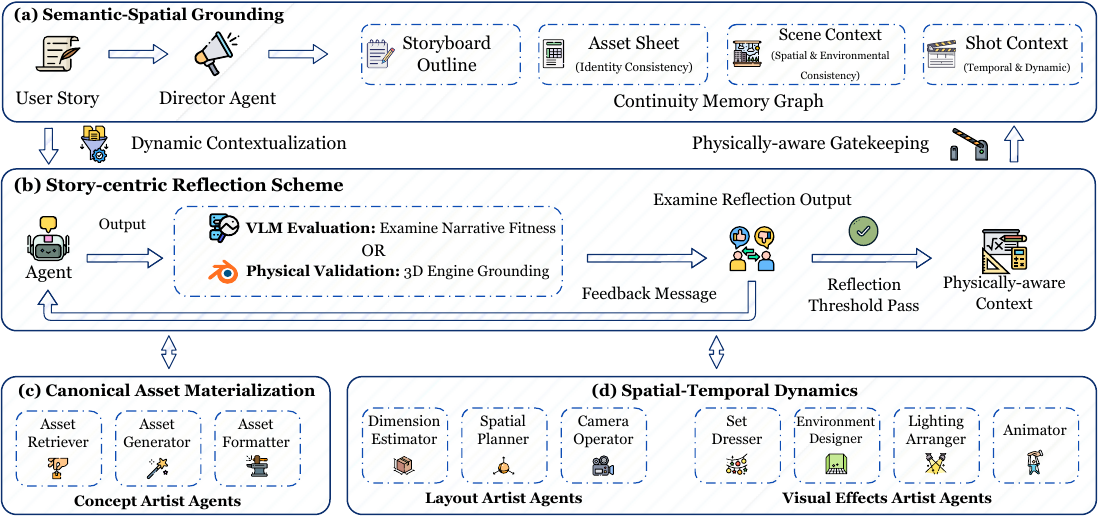}
  \captionof{figure}{
  \textbf{Hierarchical Multi-Agent Planning Framework.} 
  Governed by a \textbf{Story-centric Reflection Scheme} (b), our system utilizes iterative feedback from 3D engines, such as Blender, and Vision-Language Models to ensure geometric and narrative consistency. We translate narrative $\mathcal{T}_{story}$ into 3D storyboards $\mathcal{V}_{3D}$ via a three-stage pipeline: 
  (a) \textbf{Semantic-Spatial Grounding}, where the Director Agent decomposes the story into a structured continuity memory graph ($\bm{\mathcal{G}_{cm}}$) to ensure precise information flow to downstream agents; 
  (c) \textbf{Canonical Asset Materialization}, which instantiates entities from $\bm{\mathcal{G}_{cm}}$ to maintain global asset consistency; and 
  (d) \textbf{Spatial-Temporal Dynamics}, which performs spatial layout of assets from memory and enhances cinematic visual effects. (Details of all agents are provided in \textbf{Appendix}.)
  } \label{fig_overview1}
\end{figure}

We formulate the dynamic 3D storyboard generation task as a mapping process $F: \mathcal{T}_{story} \to \mathcal{V}_{3D}$, where a raw textual narrative script $\mathcal{T}_{story}$ is synthesized into a spatiotemporally coherent and editable 3D dynamic video sequence $\mathcal{V}_{3D}$. To bridge the gap between abstract semantics and precise spatial execution, we propose the Hierarchical Multi-Agent Planning Framework, illustrated in Fig.~\ref{fig_overview1}. This framework mimics professional film production workflows by decoupling the production pipeline into specialized roles, ensuring that each stage, from initial planning to final rendering, is handled with domain-specific expertise.
Central to this framework is the Story-centric Reflection Scheme, which shifts the generative process from a conventional feed-forward pass to an iterative optimization problem. Rather than producing a single-shot output, our agents operate within a closed-loop verification system. By executing actions within a 3D engine and verifying the results via multi-modal feedback, the system systematically mitigates spatial randomness and ensures that the final output $\mathcal{V}_{3D}$ remains faithful to the original narrative intent $\mathcal{T}_{story}$.

For any subagent $a \in \mathcal{A}_{agents}$ operating on a specific task context $\mathcal{C} \subset \bm{\mathcal{G}_{cm}}$, the initial state is generated as $o_0 = \Phi_a(\mathcal{C})$, where $\Phi_a$ denotes the generation policy of agent $a$. To prevent spatial hallucinations and ensure physical plausibility, $o_t$ is evaluated iteratively at each step $t$ by a reflection scoring function $R(o_t, \mathcal{C}) \in \mathbb{R}$, which yields diagnostic feedback $\mathcal{F}_t$:
\begin{equation}
\label{eq_reflect}
    o_{t+1} = \begin{cases} 
    o_t, & \text{if } R(o_t, \mathcal{C}) \geq \tau_a \\
    \Psi_a(o_t, \mathcal{F}_t, \mathcal{C}), & \text{if } R(o_t, \mathcal{C}) < \tau_a \text{ and } t < T_{\max} \\
    \Omega(a_{fb}, \mathcal{C}), & \text{otherwise}.
    \end{cases}
\end{equation}
Here, $\tau_a$ represents the task-specific acceptance threshold, and $T_{\max}$ is the maximum allowable reflection horizon to prevent non-convergent loops. If the criteria are not met within $T_{\max}$, the system triggers $\Omega$, a handover operator that delegates the context to a designated fallback agent $a_{fb}$ (e.g., reverting from retrieval to generative synthesis).
Depending on the subagent's domain, the feedback $\mathcal{F}_t$ and the scoring function $R$ are derived from either engine constraints (e.g., deterministic 3D engine detailing bounding box collisions) or aesthetic critiques (e.g., VLM-driven semantic alignment scores scaled to inform $\tau_a$). Driven by $\mathcal{F}_t$, the refinement operator $\Psi_a$ iteratively corrects the spatial or visual effects until the geometric and narrative requirements are satisfied.

Guided by this unified scheme, our framework is organized into three progressive modules, which are detailed in the subsequent subsections: 1) Semantic-Spatial Grounding decomposes the story into a structured continuity memory graph, enabling precise and reliable information flow to downstream agents;
2) Canonical Asset Materialization instantiates entities from the memory graph, ensuring global asset consistency;
3) Spatial-Temporal Dynamics arranges assets in space and time based on memory and enhances cinematic visual effects.

\subsection{Semantic-Spatial Grounding}
The Director Agent ($a_{dir}$) serves as the high-level orchestrator, responsible for mapping unstructured text into a structured, persistent representation: $a_{dir}: \mathcal{T}_{story} \to \bm{\mathcal{G}_{cm}}$. To systematically prevent semantic degradation over long-horizon generation, $a_{dir}$ conducts this mapping through three core mechanisms.

\mypara{Continuity Memory Graph}
Instead of relying on the transient context windows of LLMs, the director agent organizes the narrative into the continuity memory graph ($\bm{\mathcal{G}_{cm}}$). This graph acts as a hierarchical state comprising four distinct layers: (1) \textbf{Storyboard Outline} ($\mathcal{M}_{outline}$) for the temporal sequencing of scenes and shots; (2) \textbf{Asset Sheet} ($\mathcal{M}_{asset}$) for registering globally unique entity identifiers; (3) \textbf{Scene Context} ($\mathcal{M}_{scene}$) for maintaining static environmental states (e.g., layouts $\bm{L}_{layout}$ and sets $\bm{E}_{env}$); and (4) \textbf{Shot Context} ($\mathcal{M}_{shot}$) for localizing dynamic camera parameters and actions. This structure establishes a definitive source of truth by decoupling global assets from shot-specific variables.

\mypara{Dynamic Contextualization}
To actively manage information flow and prevent cognitive overload for downstream subagents, the Director employs a dynamic contextualization mechanism. We define a projection operator $\Pi_{task}(\bm{\mathcal{G}_{cm}})$ that extracts only the task-relevant subgraphs for a specific agent. Rather than exposing the entire global state, $\Pi_{task}$ supplies precise, limited constraints (e.g., providing only the active bounding boxes from $\mathcal{M}_{scene}$ to the Layout Artist Agents) while masking unrelated narrative data. By grounding subagents within a minimally sufficient working memory, this targeted routing drastically curtails semantic noise and mitigates spatial hallucinations.

\mypara{Physically-Aware Gatekeeping}
$\bm{\mathcal{G}_{cm}}$ is not a static repository but a self-evolving world representation governed by constraint-based gatekeeping. To safely integrate new agent actions, the Director utilizes a strict validation function $V(o_{new}, \mathcal{S}_{exist}) \to \{0, 1\}$. Any proposed 3D output $o_{new}$ must pass programmatic, engine-verified evaluations against the existing global state $\mathcal{S}_{exist}$ before it is committed to memory. For instance, a newly generated layout must be rigorously verified to avoid bounding box collisions. Invalid updates ($V=0$) are rejected and routed back to the offending agent's reflection loop (Eq.~\ref{eq_reflect}), ensuring long-horizon geometric consistency across the storyboard.

\subsection{Canonical Asset Materialization}

The goal of asset materialization is to instantiate abstract semantic entities from the memory graph into standardized, physical 3D meshes. Serving as the semantic-to-geometric bridge, the Concept Artist Agents ($a_{concept}$) populate the Asset Sheet ($\mathcal{M}_{asset}$) with canonical 3D geometries ($\Omega_{3D}$):
\begin{equation}
    a_{concept}: (\mathcal{M}_{outline}, \mathcal{M}_{asset}) \to \Omega_{3D}.
\end{equation}
To avoid unpredictable artifacts from direct 3D generation and to guarantee unified spatial alignment, $a_{concept}$ executes this process via a cascaded workflow as follows.

\mypara{Asset Retriever}
To mitigate topological errors, the Retriever first queries high-quality existing assets using attributes from $\mathcal{M}_{asset}$. Following Eq.~\ref{eq_reflect}, retrieved candidates $o_t$ are evaluated by a VLM reflection function $R(o_t, \mathcal{M}_{asset})$ for spatial quality and stylistic alignment with $\mathcal{M}_{outline}$. Candidates exceeding the threshold $\tau_{ret}$ are selected, ensuring optimal fidelity for standard objects without generative hallucinations.

\mypara{Asset Generator}
If retrieval fails to meet $\tau_{ret}$ within $T_{\max}$ attempts, the system triggers a \textit{Handover} ($\Omega$) to the Asset Generator. This fallback employs a two-stage synthesis: it first generates a VLM-refined 2D reference image conditioned on $\mathcal{M}_{asset}$, which is subsequently lifted into a 3D mesh. Anchoring 3D generation on a strictly verified 2D intermediate localizes geometric errors and enforces aesthetic constraints for novel assets.

\mypara{Asset Formatter}
Raw assets often reside in arbitrary local coordinate systems with unaligned forward vectors, which can cause downstream layout failures. To standardize them, the Formatter queries a VLM to identify the semantic canonical front and physical extents. It computes a transformation tuple $(\bm{R}_{align}, s_{norm})$, applying rotation $\bm{R}_{align} \in SO(3)$ for global axis alignment and uniform scaling $s_{norm}$ for bounding box normalization. This ensures all entities in $\Omega_{3D}$ share a unified mathematical space, enabling precise layout operations without orientation ambiguity. 

\subsection{Spatial-Temporal Dynamics}

The final stage of our framework transforms the canonical 3D assets into a cohesive cinematic scene. It resolves both the physical arrangement of objects (Spatial) and their narrative evolution through camera motion and character actions (Temporal), yielding the final 3D storyboard $\mathcal{V}_{3D}$.

\mypara{Layout Artist Agents}
Acting as the spatial architect, the Layout Artist Agents ($a_{layout}$) resolve scene geometry and cinematic configurations:
\begin{equation}
    a_{layout}: (\Omega_{3D}, \mathcal{M}_{scene}, \mathcal{M}_{shot}) \to (\bm{L}_{layout}, \bm{C}_{cam}).
\end{equation}
To achieve this, the agent orchestrates a three-step pipeline. First, to address the lack of scene-specific proportions in isolated assets, the \textbf{Dimension Estimator} predicts realistic scaling factors $\bm{s} \in \mathbb{R}^3$ based on semantic roles. This ensures physically plausible relative sizes across the environment. 
Second, to bridge abstract text and precise 3D coordinates, the \textbf{Spatial Planner} maps qualitative prepositions into quantitative transformations (translation $\bm{t} \in \mathbb{R}^3$, rotation $\bm{r} \in SO(3)$) to form the layout $\bm{L}_{layout}$. The 3D engine programmatically verifies this proposed layout, detecting physical violations such as collisions. By returning exact error logs to the reflection loop, the planner iteratively corrects the arrangement, ensuring geometric validity.

Finally, to translate the static layout into a cinematic experience, the \textbf{Camera Operator} determines the camera state $\bm{C}_{cam}$ via VLM-guided Discrete Semantic Visual Servoing. To ensure initial visibility, the camera pivot initializes at the target's bounding box centroid with an orbital distance geometrically derived to guarantee a specific field of view (FOV). To dynamically refine this framing, the VLM evaluates rendered frames and issues discrete camera-control commands (e.g., \textit{Orbit, Pan}) mapped to differential spatial updates. This closed-loop servoing optimizes the semantic alignment between the rendered view and the directorial intent. (Details are provided in \textbf{Appendix}.)

\mypara{Visual Effects Artist Agents}
To elevate the geometrically valid layout into a fully realized cinematic sequence, the Visual Effects (VFX) Artist Agents ($a_{vfx}$) finalize the atmospheric and temporal dimensions:
\begin{equation}
    a_{vfx}: (\mathcal{M}_{scene}, \mathcal{M}_{shot}) \to (\bm{I}_{light}, \bm{E}_{env}, \bm{P}_{action}).
\end{equation}
To achieve this, the agent executes a four-step enhancement pipeline. First, to spatially ground the isolated primary assets, the \textbf{Environment Designer} constructs structural backdrops ($\bm{E}_{env}$). This establishes clear spatial boundaries and contextual realism for the scene. Second, to overcome the visual sparsity of the initial layout, the \textbf{Set Dresser} dynamically places supplementary props without violating existing bounding box constraints. This targeted dressing enriches fine-grained scene detail and enhances overall visual plausibility.

Third, to drive the cinematic storytelling, the \textbf{Lighting Arranger} configures environmental lighting ($\bm{I}_{light}$) using suitable high dynamic range image (HDRI) maps and discrete light sources. By iteratively refining light intensity and color via VLM-guided aesthetic reflection, it ensures precise emotional alignment with the script's intended mood. Finally, to transition the scene from static frames to a dynamic narrative, the \textbf{Animator} retrieves appropriate motion sequences from an animation database and retargets them to the characters ($\bm{P}_{action}$). This comprehensive pipeline successfully drives the temporal evolution of the storyboard, ultimately yielding a rich, physically grounded, and narratively expressive 3D video sequence. (Details are provided in \textbf{Appendix}.)

\section{Experiments}

\subsection{Implementation Details}

\mypara{Baselines}
We evaluate StoryBlender against two groups of baselines representing distinct generation schemes. The first group consists of leading 2D generative models and AI storyboarding tools (AniMaker~\cite{shi2025AniMaker}, Qwen-Edit~\cite{wu2025qwen}, Story2Board~\cite{dinkevich2025Story2Board}, StoryDiffusion~\cite{zhou2024storydiffusion}), which serve as representatives of state-of-the-art pixel-space visual storytelling. The second group comprises contemporary 3D scene generation and reconstruction frameworks (SceneWeaver~\cite{yang2025sceneweaver} and a cascaded pipeline consisting of Qwen-Edit~\cite{wu2025qwen} for initial image generation followed by CAST~\cite{yao2025cast} for 3D reconstruction), which serve as representative methods for isolated 3D scene generation and post-hoc 2D-to-3D reconstruction.

\begin{figure}[ht]
    \centering
    \includegraphics[width=\linewidth]{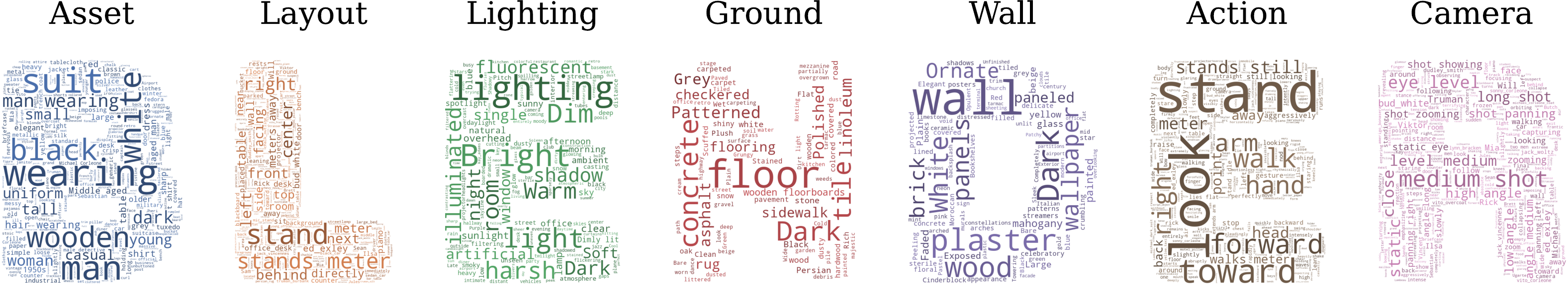}
    \caption{\textbf{Narrative Concept Visualization of CineBoard3D.}} 
    \label{fig:wordclouds}
\end{figure}

\mypara{Dataset}
To evaluate our framework in a high-fidelity filmmaking context, we introduce \textbf{CineBoard3D}, a novel dataset curated from eight iconic feature films, such as \textit{The Godfather} and \textit{Pulp Fiction}. CineBoard3D comprises 51 scenes and 178 shots, featuring 95 characters and 81 props. With an average of 22.25 shots per story and rigorous continuity requirements across diverse environments, it provides a significantly more complex and robust testbed for multi-shot spatial consistency than existing text-to-visual benchmarks. Fig.~\ref{fig:wordclouds} visualizes the word clouds for the asset, layout, environment, and action descriptions found within the dataset.

\mypara{Evaluation Metrics}
We evaluate our framework using three key metric categories. \textbf{Consistency Metrics} measure sequence stability across two dimensions: \textbf{Self} evaluates intra-sequence coherence (consistency among the generated shots), while \textbf{Cross} measures reference-anchored fidelity (adherence to the initial global asset). We apply these dimensions to both \textbf{Character Identification Similarity (CIDS)}, which computes the cosine similarity of character features, and \textbf{CLIP Style Disentanglement (CSD)}, which uses a style-trained CLIP encoder to assess global aesthetic coherence. \textbf{Onstage Character Count Matching (OCCM)} calculates the numerical accuracy of generated characters against script requirements to penalize spatial hallucinations or omissions. Finally, \textbf{Prompt Alignment (PA)} employs a large multimodal model (GPT-4.1) as a judge to rate generated images on a $0$--$4$ scale, assessing both scene alignment (environment/layout) and action alignment (character interactions).

\mypara{Setup}
We conduct our experiments on the proposed CineBoard3D dataset. Our multi-agent framework employs Gemini 3 Pro~\cite{team2023gemini} for high-level reasoning and Gemini 3 Flash to drive the remaining subagents, and the iterative Story-centric Reflection loop, which is capped at a maximum of $5$ iterations, with the reflection threshold $\tau_a$ set to $8$. Scene population utilizes assets retrieved from Poly Haven and Sketchfab. For assets, we use Nano Banana~\cite{team2023gemini} for text-to-image synthesis, followed by Hunyuan 3D~\cite{lai2025hunyuan3d} for image-to-3D generation. The end-to-end pipeline is integrated within Blender via the Model Context Protocol.

\begin{table}[t]
\centering

\caption{
\textbf{Quantitative comparison on the CineBoard3D dataset.} We evaluate performance across multiple dimensions: Character Identification Similarity (CIDS), Onstage Character Count Matching (OCCM), Style Similarity (CSD), Prompt Alignment (PA), and User Study (rank 1--7, Details in the \textbf{Appendix}.). 
\textbf{Bold} and \underline{underlined} indicate the best and second-best results.
}
\label{tab:main_results}

\setlength{\tabcolsep}{6pt}
\resizebox{0.95\textwidth}{!}{%
\renewcommand{\arraystretch}{1.25} 
\begin{tabular}{@{} l c c c c c c c c @{}} 
\toprule

\multirow{2}{*}[-0.5ex]{\textbf{Method}} & 
\multicolumn{2}{c}{\textbf{CIDS} $\uparrow$} & 
\textbf{OCCM} $\uparrow$ & 
\multicolumn{2}{c}{\textbf{CSD} $\uparrow$} & 
\multicolumn{2}{c}{\textbf{PA} $\uparrow$} & \textbf{User $\downarrow$} \\

\cmidrule(lr){2-9}

 & \textbf{Self} & \textbf{Cross} & \textbf{Shots} & \textbf{Self} & \textbf{Cross} & \textbf{Scene} & \textbf{Action} & \textbf{Rank}\\
\midrule

AniMaker & $0.801$ & $0.613$ & \underline{$73.91$} & $0.561$ & $0.309$ & \underline{$3.689$} & \underline{$3.369$} & $3.46$ \\
Qwen-Edit & $0.849$ & $0.719$ & $67.94$ & $0.503$ & $0.296$ & $\bm{3.736}$ & $\bm{3.613}$ & \underline{$2.23$} \\
Story2Board & $0.842$ & $0.772$ & $49.30$ & $0.513$ & $0.264$ & $3.189$ & $2.281$ & $5.25$ \\
StoryDiffusion & $0.816$ & $0.758$ & $62.89$ & $0.612$ & \underline{$0.343$} & $2.547$ & $1.815$ & $5.80$ \\
Qwen-Edit+CAST & $0.726$ & $0.631$ & $56.39$ & $0.727$ & $0.262$ & $0.800$ & $1.866$ & $6.58$ \\
SceneWeaver & \underline{$0.872$} & \underline{$0.773$} & $55.66$ & \underline{$0.758$} & $0.216$ & $2.824$ & $1.933$ & $4.13$ \\

\midrule
\textbf{StoryBlender} & $\bm{0.917}$ & $\bm{0.803}$ & $\bm{78.16}$ & $\bm{0.772}$ & $\bm{0.352}$ & $3.176$ & $2.629$ & $\bm{1.84}$ \\

\bottomrule
\end{tabular}%
}
\end{table}

\subsection{Main Results}
\mypara{Quantitative Comparisons}
StoryBlender's reliance on a deterministic 3D world model resolves the stability and consistency issues in pixel-based generative baselines, yielding stronger quantitative performance across key metrics. Specifically, the framework achieves \textbf{better identity preservation}, scoring $0.917$ in CIDS (Self) and $0.803$ in CIDS (Cross), demonstrating that the continuity memory graph anchors identities to persistent 3D meshes rather than relying on pixel-space mechanisms for multi-view consistency, as in StoryDiffusion or Story2Board. Compared with single-shot 3D generation methods like SceneWeaver, StoryBlender shows greater consistency across these metrics, validating the utility of the memory graph in maintaining coherence across multiple cuts. This advantage also extends to global lighting and atmospheric stability, reflected in a CSD score of $0.772$. Furthermore, StoryBlender mitigates the common failure mode of hallucinating or omitting characters, achieving the highest \textbf{Onstage Character Count Matching} with a score of $78.16$ by instantiating discrete models from the scene registry rather than relying on pixel probabilities. While pixel-centric methods like Qwen-Edit score higher in \textbf{Prompt Alignment} by bending physical constraints to satisfy prompts, StoryBlender trades this freedom for physical plausibility, prioritizing spatial validity over semantic compliance. 
These metrics and the user study on asset consistency, environmental consistency, and prompt adherence show that StoryBlender balances narrative requirements with geometric constraints for consistent, spatially coherent storyboards.

\begin{figure}[!ht]
    \centering
    \includegraphics[width=\textwidth]{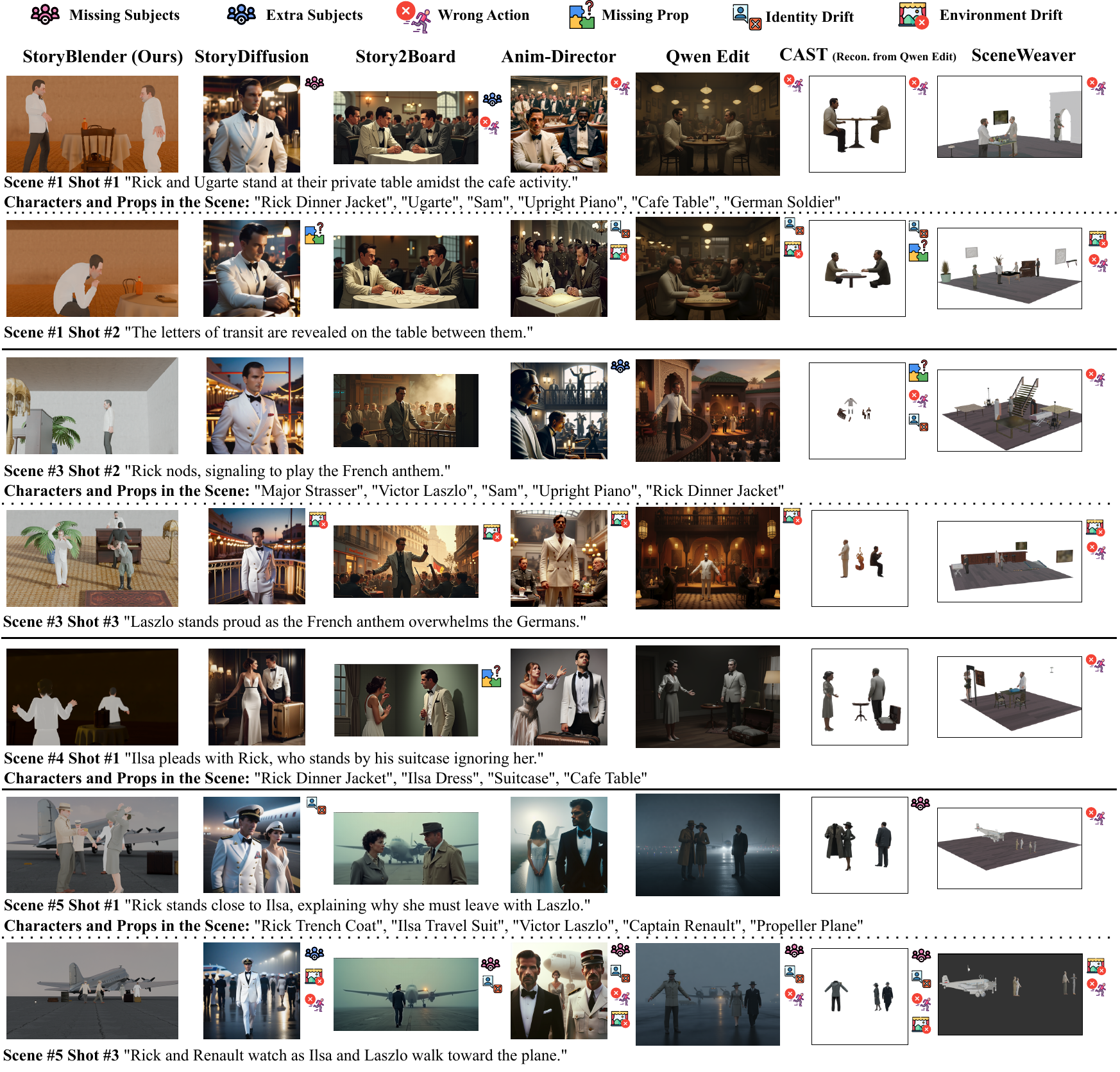}
    \caption{\textbf{Comparison of baselines on a complex multi-shot sequence from the film \textit{Casablanca}.} StoryBlender demonstrates stronger geometric consistency and entity management across shots, maintaining the architectural layout and correct character count in each frame. In contrast, StoryDiffusion and Story2Board capture the general semantic atmosphere but exhibit spatial inconsistencies, hallucinating background changes between camera cuts, and failing to preserve the correct number of characters. (More stories and shots are presented in the \textbf{Appendix}.)}
    \label{fig:qualitative}
\end{figure}

\mypara{Qualitative Analysis}
As illustrated by the multi-shot case study of \textit{Casablanca} (see Fig.~\ref{fig:qualitative}), StoryBlender outperforms 2D visual narrative baselines by prioritizing physical logic over semantic approximation. Unlike pixel-based models that suffer from environmental drift and hallucinate background changes between camera cuts, StoryBlender achieves \textbf{geometric stability} by anchoring architectural features and spatial constraints within the $\mathcal{M}_{scene}$ layer, keeping them consistent regardless of camera trajectory. Additionally, the framework executes \textbf{precise spatial blocking} by translating semantic prepositions into Euclidean transformations, allowing it to resolve constraints like depth ordering and occlusion that 2D methods flatten or distort. This approach also ensures \textbf{reliable numerical and entity management}; by instantiating the required number of 3D meshes cataloged in the $\mathcal{M}_{asset}$ registry, the system addresses the ``vanishing'' or ``multiplying'' entity problems common in probabilistic generation, reducing unprompted additions or omissions. Consequently, by maintaining entity fidelity, spatial execution, and environmental stability across dynamic shots, StoryBlender provides a dependable and geometrically sound foundation that surpasses traditional generative baselines.

\begin{table}[t]
\centering
\begin{minipage}[t]{0.49\linewidth}
\centering
\caption{\textbf{Ablation of the continuity memory graph on the CineBoard3D.}}
\label{tab:ablation_cmg}
\renewcommand{\arraystretch}{1.2}
\setlength{\tabcolsep}{2pt}
\resizebox{\linewidth}{!}{%
\begin{tabular}{@{}lccc@{}}
\toprule
\textbf{Configuration} & \textbf{CIDS (Self)} $\uparrow$ & \textbf{CIDS (Cross)} $\uparrow$ & \textbf{SDE} $\downarrow$ \\ \midrule
w/o $\bm{\mathcal{G}_{cm}}$ & $0.547$ & $0.382$ & $4.87$ \\
w/o $\mathcal{M}_{asset}$ & $0.532$ & $0.414$ & $0.95$ \\
w/o $\bm{L}_{layout}$ & $0.908$ & $0.788$ & $4.12$ \\ \midrule
\textbf{Full $\bm{\mathcal{G}_{cm}}$ (Ours)} & $\bm{0.917}$ & $\bm{0.803}$ & $\bm{0.28}$ \\ \bottomrule
\end{tabular}%
}
\end{minipage}\hfill
\begin{minipage}[t]{0.47\linewidth}
\centering
\caption{\textbf{Ablation of Reflection and Visual Effects Agents.}}
\label{tab:ablation_pa}
\renewcommand{\arraystretch}{1.2}
\setlength{\tabcolsep}{2pt}
\resizebox{\linewidth}{!}{%
\begin{tabular}{@{}lcc@{}}
\toprule
\textbf{Configuration} & \textbf{PA (Scene)} $\uparrow$ & \textbf{PA (Action)} $\uparrow$ \\ \midrule
w/o Visual Effects Agents & $0.561$ & $1.374$ \\
w/o Reflection ($t=0$) & $2.867$ & $2.540$ \\
w/ Reflection ($t=3$) & $3.114$ & $2.621$ \\ \midrule
\textbf{Full Pipeline (Ours)} & $\bm{3.176}$ & $\bm{2.629}$ \\ \bottomrule
\end{tabular}%
}
\end{minipage}
\end{table}

\subsection{Ablation Study}

\mypara{Semantic-Spatial Grounding} 
We evaluate the role of the hierarchical state machine in maintaining multi-shot continuity by ablating the continuity memory graph. We compare four configurations: (1) \textbf{w/o $\bm{\mathcal{G}_{cm}}$}: a pure LLM baseline processing shot narratives independently; (2) \textbf{w/o $\mathcal{M}_{asset}$}: shares the scene layout but lacks a persistent asset registry; (3) \textbf{w/o $\bm{L}_{layout}$ in $\mathcal{M}_{scene}$}: uses the global asset registry for mesh consistency but computes spatial layout $\bm{L}_{layout}$ per shot; and (4) \textbf{Full $\bm{\mathcal{G}_{cm}}$}. To quantify geometric instability, we introduce \textbf{Spatial Drift Error (SDE)}, measuring the average Euclidean distance of the 8 world-space bounding box corners $v_{j,i}$ for static assets $j \in \mathcal{S}_{static}$ across $K$ consecutive shots:
\begin{equation}
\text{SDE} = \frac{1}{|\mathcal{S}_{static}|} \sum_{j \in \mathcal{S}_{static}} \frac{1}{K-1} \sum_{k=1}^{K-1} \left( \frac{1}{8} \sum_{i=1}^{8} \| v_{j,i}^{(k+1)} - v_{j,i}^{(k)} \|_2 \right).
\end{equation}
As shown in Table~\ref{tab:ablation_cmg}, removing $\bm{\mathcal{G}_{cm}}$ causes severe identity and geometric failures, indicating that standard LLMs operating over natural-language context windows struggle to maintain multi-shot continuity. Without the asset registry (\textbf{w/o $\mathcal{M}_{asset}$}), CIDS drops significantly, showing that text prompts alone cannot prevent 3D mesh identity drift. Omitting the shared layout (\textbf{w/o $\bm{L}_{layout}$ in $\mathcal{M}_{scene}$}) preserves identity but increases SDE to $4.12$, revealing spatial inconsistency where background objects shift between shots without persistent spatial context. The full $\bm{\mathcal{G}_{cm}}$ resolves both issues, achieving consistent identity and stable geometry.

\begin{wrapfigure}{r}{0.48\textwidth}
  \centering
  \vspace{-10pt} 
  \includegraphics[width=\linewidth]{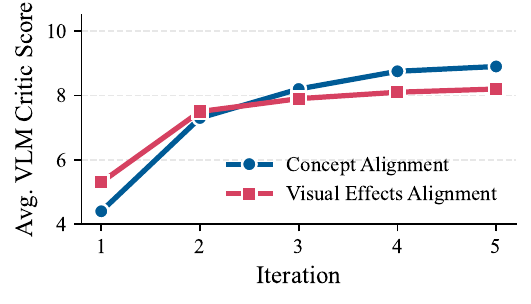}
  \caption{
  \textbf{Impact of Story-centric Reflection for the Concept Artist and Visual Effects Artist Agents.}
  }
  \label{fig:ablation_reflection_vlm}
  \vspace{-10pt} 
\end{wrapfigure}
\mypara{Visual Effects Artist Agents} 
To assess the impact of specialized roles, we remove the Visual Effects Artist Agents from the pipeline. We replace the atmospheric lighting and character animations with a default Nishita sky texture~\cite{nishita1993display} and static rest poses, causing Prompt Alignment scores to drop (see Table~\ref{tab:ablation_pa}). This highlights that geometric layout alone is insufficient. Specialized subagents are required to translate semantic mood and motion into perceptible visual signals. (Visualization cases in the \textbf{Appendix}.)

\begin{figure}[t]
  \centering

  \begin{minipage}[c]{0.48\linewidth}
    \centering
    \captionof{table}{\textbf{Spatial error over five reflection turns for naive and physical methods.} D: Direction, R: Relationship, O: Occlusion, C: Contact.}
    \label{tab:layout_errors}
    \renewcommand{\arraystretch}{1.2}
    \setlength{\tabcolsep}{2pt}
    \resizebox{0.95\linewidth}{!}{%
    \begin{tabular}{@{}ccccccc@{}}
    \toprule
    \multirow{2}{*}{\textbf{Method}} & \multirow{2}{*}{\textbf{Error}} & \multicolumn{5}{c}{\textbf{Turn}} \\ 
    \cmidrule(l){3-7}
     &  & \textbf{1} & \textbf{2} & \textbf{3} & \textbf{4} & \textbf{5} \\ 
    \midrule
    
    \multirow{4}{*}{Naive} 
    & D & 0.341 & 0.212 & 0.188 & 0.129 & 0.129 \\
    & R & 0.224 & 0.129 & 0.141 & 0.094 & 0.094 \\
    & O & 0.012 & 0.000 & 0.012 & 0.000 & 0.012 \\
    & C & 0.000 & 0.012 & 0.000 & 0.000 & 0.000 \\
    
    \midrule
    
    \multirow{4}{*}{Physical} 
    & D & 0.377 & 0.049 & 0.012 & 0.008 & 0.008 \\
    & R & 0.224 & 0.110 & 0.000 & 0.000 & 0.000 \\
    & O & 0.041 & 0.012 & 0.000 & 0.004 & 0.000 \\
    & C & 0.008 & 0.008 & 0.000 & 0.000 & 0.000 \\
    
    \bottomrule
    \end{tabular}
    }
  \end{minipage}
  \hfill
\begin{minipage}[c]{0.5\linewidth}
    \centering
    \vspace{15pt}
    \includegraphics[width=\linewidth]{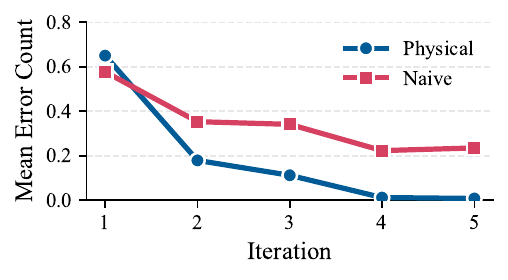}
    \caption{\textbf{Impact of Story-centric Reflection for the Layout Artist Agents.} Compared to a naive reflection approach, 3D physical reflection rapidly minimizes spatial errors across reflection turns.}
    \label{fig:ablation_reflection_engine}
  \end{minipage}
\end{figure}

\mypara{Story-centric Reflection}
We analyze the efficacy of the iterative feedback loop by halting generation at $t=0$ and $t=3$. Removing this refinement ($t=0$) causes Prompt Alignment scores to regress (see Table~\ref{tab:ablation_pa}), indicating that zero-shot generation often misses fine-grained constraints. Permitting a partial reflection horizon ($t=3$) notably improves the scores to $3.114$ (Scene) and $2.621$ (Action), demonstrating that intermediate feedback effectively corrects errors, though the full five-turn pipeline yields the best overall alignment. Fig.~\ref{fig:ablation_reflection_vlm} visualizes the system's recovery, plotting the consistent upward trajectory of VLM critic scores for Concept and Visual Effects Alignment across five iterations to demonstrate semantic convergence. 
To evaluate the Layout Artist's spatial reasoning, we compare our engine-driven physical reflection against a naive reflection baseline, which tasks the LLM with fixing previous errors without any spatial context from the 3D engine. As illustrated in Fig.~\ref{fig:ablation_reflection_engine}, the physical reflection achieves a significantly faster error reduction rate. Table~\ref{tab:layout_errors} quantifies this geometric self-correction: relying on engine feedback, spatial errors (direction, relationship, contact, and occlusion) drop substantially after the first iteration and reach near-zero by the third verification turn. In contrast, the naive approach plateaus, confirming that LLMs struggle to resolve complex layout constraints without grounded 3D state feedback. (Visualization cases in the \textbf{Appendix}.)

\begin{figure}[t]
  \centering
  \includegraphics[width=1.0\linewidth]{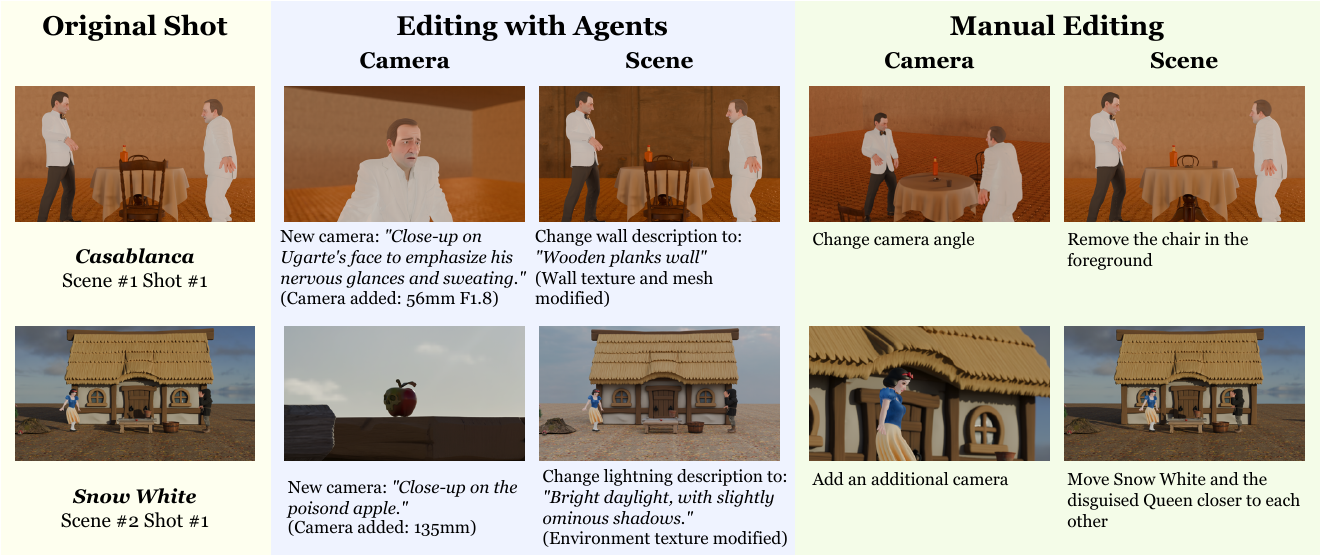} 
    \caption{
    \textbf{Non-destructive editing workflow.} Starting from an initial generation (left), the figure shows two modes of modification. \textbf{Middle:} agent-assisted edits, where natural language updates camera placement, lighting, and textures without changing unrelated geometry. \textbf{Right:} manual edits in the 3D engine for asset addition/removal, asset transformation, and camera adjustment. This dual-mode design demonstrates precise editability.
}
\label{fig:edit}
\end{figure}

\subsection{In-Place Scene Editing}
A limitation of 2D generative pipelines is their destructive editing process: modifying an element (e.g., camera angle) typically requires re-sampling the entire image, often removing existing details. In contrast, StoryBlender produces a parametric 3D scene graph that supports non-destructive edits through a dual-mode workflow, combining agent-assisted updates and manual editing (Fig.~\ref{fig:edit}).
Scene attributes can be modified through natural language instructions. For example, ``switch to a close-up shot'' updates camera extrinsics ($\bm{C}_{cam}$) while preserving geometry. Similarly, instructions such as ``change the lighting to midday'' or ``replace the concrete wall with wood'' update lighting HDRI ($\bm{I}_{light}$) and backdrop textures ($\bm{E}_{env}$). These edits remain localized and do not affect other elements.
For finer control, StoryBlender exposes the native 3D project file for direct scene graph editing. Users can add or remove props and adjust camera parameters directly in the engine, preserving geometric consistency and enabling the storyboard to function as an editable production asset. 
This capability allows iterative adjustments without regenerating the scene, improving practical usability for storyboard design.

\section{Conclusion}

\begin{figure}[ht]
    \centering
    \includegraphics[width=\linewidth]{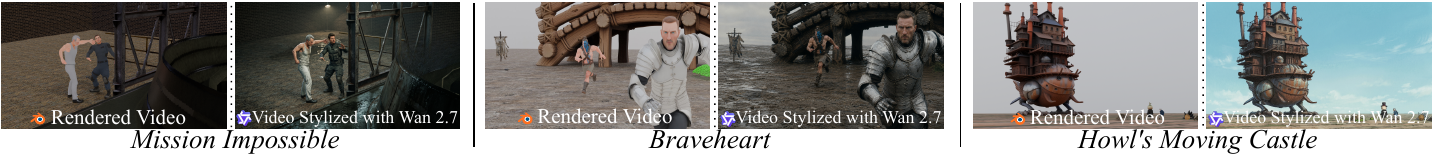}
    \caption{\textbf{Videos stylized with Wan 2.7.}}
    \label{fig:stylize_with_wan}
\end{figure}

We introduced StoryBlender, a hierarchical multi-agent framework for generating consistent and editable 3D storyboards from narrative scripts. By grounding generation in a persistent continuity memory graph and a reflection-based planning loop, the system bridges semantic story understanding with physically valid spatial execution, enabling reliable coordination among narrative planning, asset reasoning, and geometric validation. Our pipeline decomposes story generation into semantic grounding, asset materialization, and spatial-temporal dynamics, enabling multi-shot consistency and explicit control within a 3D engine while providing a structured and interpretable workflow for cinematic scene construction.
Future work will extend the continuity memory graph with richer semantic and physical attributes, enabling more complex scene reasoning, such as fine-grained object interactions and long-horizon narrative dependencies; combine generated 3D storyboards with video generation models (see Fig.~\ref{fig:stylize_with_wan}), and add audio to rendered clips through video-to-audio generation~\cite{cheng2024taming,li2025tri}. We also plan to expand the agent pipeline with stronger motion modeling and interactive editing capabilities to support dynamic character behaviors, more expressive cinematography, and tighter human–AI collaboration in production workflows.

%
%
\bibliographystyle{splncs04}
\bibliography{main}
\end{document}